\documentclass[a4paper]{article}

\usepackage{INTERSPEECH2022,amsmath,graphicx,amsfonts,amssymb, bm, url, subfig}

\title{Self-Supervised Learning for Multi-Channel Neural Transducer}
\name{Atsushi Kojima}
\address{Advanced Media, Inc.}
\email{a-kojima@advanced-media.co.jp}

\begin{document}

\maketitle
\begin{abstract}
\mbox{Self-supervised} learning, such as with the wav2vec 2.0 framework significantly improves the accuracy of \mbox{end-to-end} automatic speech recognition (ASR). Wav2vec 2.0 has been applied to \mbox{single-channel} \mbox{end-to-end} ASR models. In this work, we explored a \mbox{self-supervised} learning method for a \mbox{multi-channel} \mbox{end-to-end} ASR model based on the wav2vec 2.0 framework. As the \mbox{multi-channel} \mbox{end-to-end} ASR model, we focused on a \mbox{multi-channel} neural transducer. In \mbox{pre-training}, we compared three different methods for feature quantization to train a \mbox{multi-channel} conformer audio encoder: joint quantization, \mbox{feature-wise} quantization and \mbox{channel-wise} quantization. In \mbox{fine-tuning}, we trained the \mbox{multi-channel} \mbox{conformer--transducer}. All experiments were conducted using the \mbox{far-field in-house} and \mbox{CHiME-4} datasets. The results of the experiments showed that \mbox{feature-wise} quantization was the most effective among the methods. We observed a 66\% relative reduction in character error rate compared with the model without any \mbox{pre-training} for the \mbox{far-field in-house} dataset. 

\end{abstract}


\section{Introduction}
\label{sec:intro}
\mbox{Self-supervised} learning significantly improves the accuracy of \mbox{end-to-end} automatic speech recognition (ASR) models such as the \mbox{attention-based} \mbox{encoder-decoder} \cite{las}, connectionist temporal classification (CTC) \cite{ctc} and neural transducers \cite{rnnt}. The most popular framework for \mbox{self-supervised} learning is wav2vec 2.0 \cite{w2v}. 
In wav2vec 2.0 pre-training, the model is trained similarly to that in masked language modeling \cite{bert}. In \mbox{fine-tuning}, the model is trained using the ASR loss function. Wav2vec 2.0 has been mainly applied to \mbox{single-channel end-to-end} ASR models \cite{w2v, mel2vec, w2v_c}. In this work, we train a \mbox{multi-channel} \mbox{end-to-end} ASR model based on the wav2vec 2.0 framework.

Multi-channel \mbox{end-to-end} ASR models can improve the robustness of \mbox{far-field} ASR in noisy environments, because the models can capture not only spectral information but also spatial information of the target and interference signals captured from different microphones \cite{beam, chime4}. In many \mbox{end-to-end multi-channel} ASR architectures \cite{multi_las1, multi_las2, multi_las3,multi_las4,multi_tt, bf_g}, a \mbox{multi-channel} neural transducer \cite{multi_tt} is promising in terms of efficiency and accuracy. 

\mbox{Multi-channel} neural transducers are based on neural transducers such as \mbox{Transformer--Transducer} \cite{t_t1, t_t2} and \mbox{Conformer--Transducer} \cite{conformer}. \mbox{Multi-channel} neural transducers can learn the contextual relationship across channels using \mbox{channel-wise} and \mbox{cross-channel} \mbox{self-attention} layers without beamforming \cite{mask_bf, mask_bf2, frame}. \mbox{Multi-channel} neural transducers outperform typical \mbox{multi-channel} \mbox{end-to-end} ASR models, which are cascaded with neural beamforming \cite{multi_tt}.

In this work, we train a \mbox{multi-channel} neural transducer based on wav2vec 2.0 \mbox{pre-training}. 
For training, we explore three quantization methods: joint quantization, \mbox{feature-wise} quantization and \mbox{channel-wise quantization}. 
We report the results of experiments using the \mbox{far-field in-house} and public \mbox{CHiME-4} datasets \cite{chime4}. 

In the experiments, we show that \mbox{feature-wise quantization} has the best performance among the quantization methods. We observe 66\% and 4.2\% relative reductions in character error rate compared with the model without any \mbox{pre-training} for the \mbox{far-field in-house} and \mbox{CHiME-4} datasets, respectively.

\section{Background}
\label{sec:back}
\subsection{Multi-channel neural transducer}
\label{sec:t-t}
\begin{figure}[htb]
	\centering
	\includegraphics[scale=0.33]{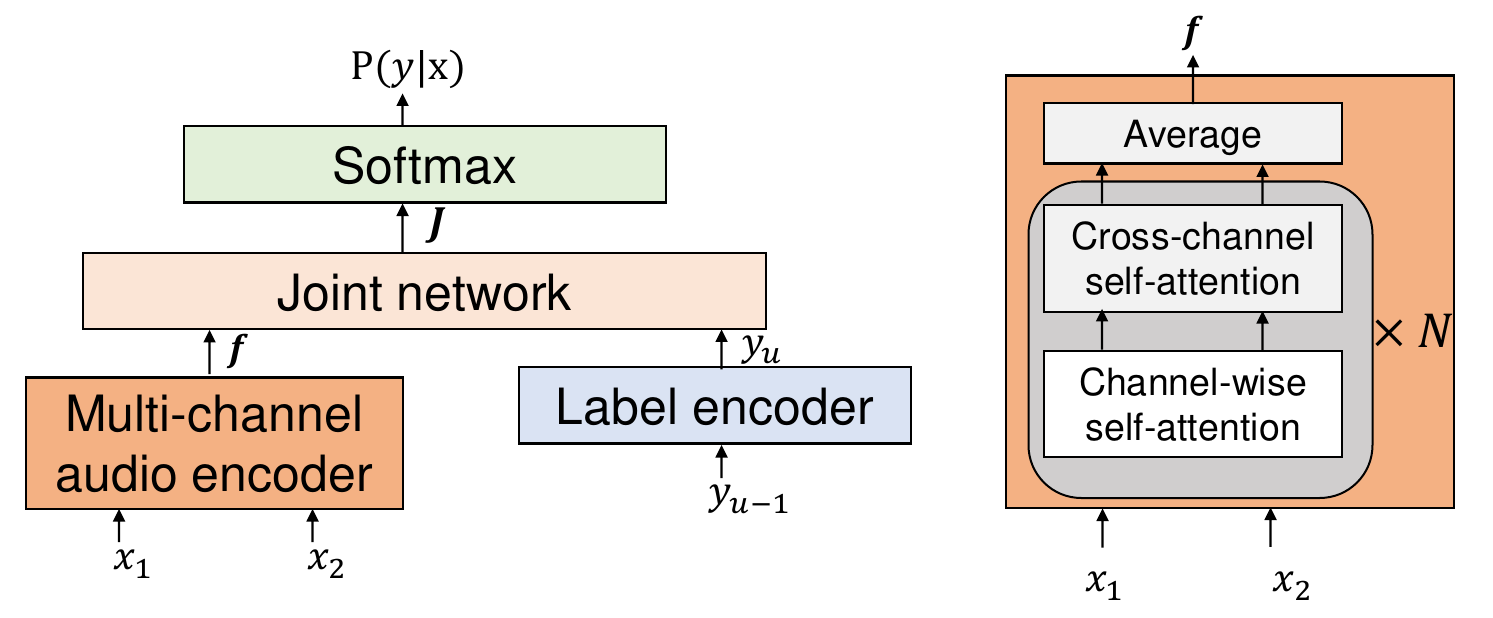}	
	\caption{Architecture of multi-channel neural transducer in the case of two channels.}
	\label{fig:transducer}
\end{figure}

We first describe the architecture of a \mbox{multi-channel} neural transducer. Figure~\ref{fig:transducer} shows an overview of a \mbox{multi-channel} neural transducer in the case of two channels. Given acoustic feature $\bm{x}$ and previous tokens $\bm{y}_{u-1}$, the \mbox{multi-channel} audio encoder converts acoustic feature $\bm{x}$ to hidden vector $\bm{f}$, and the label encoder predicts a new token $\bm{y}_U$ based on past tokens except for a blank token. The joint network outputs vector $J$ using two hidden vectors from audio and label encoders, and softmax outputs logits.

A \mbox{multi-channel} audio encoder consists of \mbox{channel-wise} \mbox{self-attention} and \mbox{cross-channel} \mbox{self-attention} layers. The \mbox{channel-wise self-attention} layers convert inputs from each channel to hidden vectors independently via multi-head attention (MHA) \cite{transformer}. The \mbox{cross-channel self-attention} layers learn the contextual relationship across channels. We convert hidden vector $h_i$ of the $i$th channel from the \mbox{channel-wise self-attention} layers to query $Q_i$, and the mean vector of the hidden vectors of other channel inputs from the \mbox{channel-wise self-attention} layers is converted to key ${K_i}$ and value ${V_i}$. The MHA is calculated using $Q_i$, $K_i$ and $V_i$. Finally, hidden vectors from the \mbox{cross-channel} \mbox{self-attention} layers are fused by taking a simple average. For input features, a \mbox{multi-channel} audio encoder obtains not only amplitude features but also phase features, unlike a \mbox{single-channel} neural transducer.

A \mbox{multi-channel} neural transducer is trained using the \mbox{recurrent neural network--Transducer} (\mbox{RNN--T}) loss \cite{rnnt}. Given acoustic features $\bm{x}$ and label sequence $\bm{y}$, the neural transducer outputs $T \times U$ logits. The \mbox{RNN--T} loss is calculated as the sum of probabilities for all paths using a \mbox{forward--backward} algorithm. The \mbox{RNN--T} loss function is written as
\begin{equation}
\label{eq:loss}
\lambda_{\rm RNN-T}=-\sum_{i}^{} \log{P(\bm{y}|{\rm x})},
\end{equation}
\begin{equation}
\label{eq:prob}
P(\bm{y}|{\rm x})=\sum_{\bm{J} \in \mathcal{Z}(\bm{y}, T)}^{} P(J|{\rm x}),
\end{equation}
where $\mathcal{Z}(\bm{y}, T)$ is the set of all alignments of length $T$ for the token sequence.

\subsection{Self-supervised learning based on wav2vec 2.0 framework}
\label{sec:contrastive_loss}
We next describe \mbox{self-supervised} learning based on the wav2vec 2.0 framework. In wav2vec 2.0 \mbox{pre-training}, the audio encoder is trained by minimizing the contrastive loss. Given target quantized feature ${\bm q}_t$ and $K$ distractors (non-target quantized features), the model must identify the true quantized feature among $K+1$ quantized features ${\bm{\tilde{q}} \in \bm{Q}_t}$. 
The contrastive loss is calculated as
\begin{equation}
\lambda=-\log{ \frac{\exp(sim(\bm{f}_t, \bm{q}_t))}{\sum_{\bm{\tilde{q}}\sim \bm{Q}_t}^{}\exp(sim(\bm{f}_t,\bm{\tilde{q}}_t))}},
\label{eq:cont_loss}
\end{equation}
where 
$\bm{f}$ denotes hidden vectors from the masked audio encoder and $sim$ is a function for calculating the cosine similarity between two vectors: $sim(\bm{a},\bm{b})=\bm{a}^T\bm{b}/||\bm{a}||||\bm{b}||$.

An audio encoder consists of Transformer \cite{transformer} or Conformer \cite{conformer}. A quantizer consists of a quantization \cite{w2v} or linear \cite{mel2vec} layer.

\section{Self-supervised learning for multi-channel neural transducer}
\label{proposed}
For the training of the \mbox{multi-channel} audio encoder, we explore three quantization methods: joint quantization, \mbox{feature-wise} quantization and \mbox{channel-wise quantization}.

\subsection{Joint quantization}
\label{proposed0}
\begin{figure}[htb]
	\centering
	\includegraphics[scale=0.31]{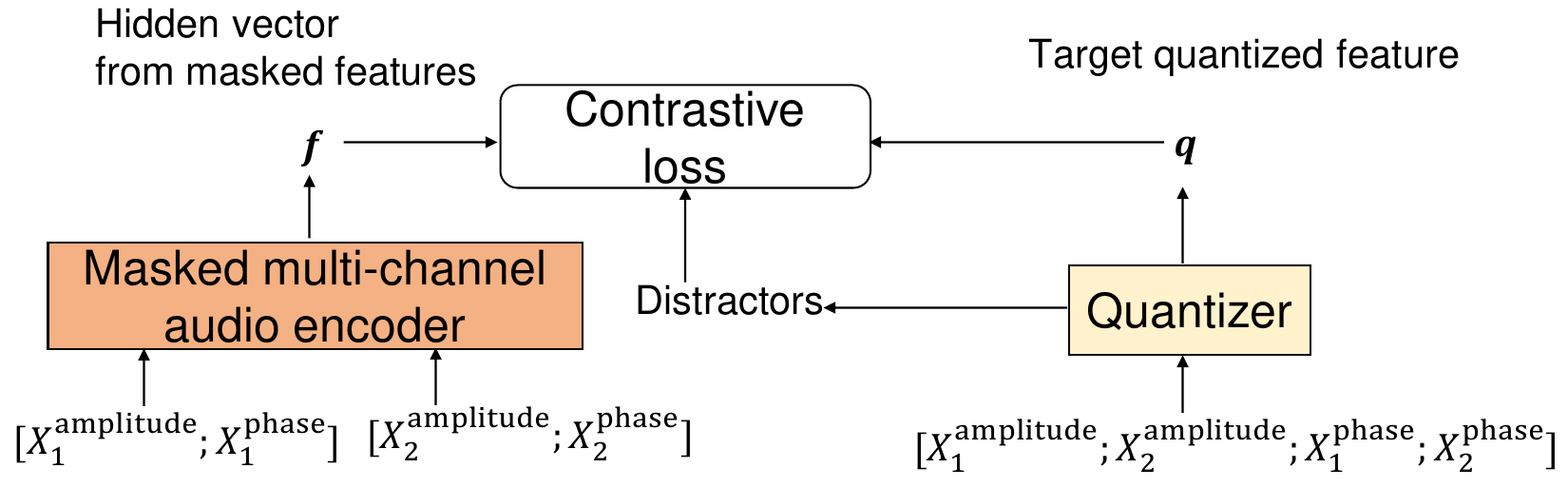}	
	\caption{Joint quantization in the case of two channels.}
	\label{fig:p1}
\end{figure}

Figure~\ref{fig:p1} shows joint quantization. In this figure, $X^{\rm amplitude}$ and $X^{\rm phase}$ 
are the amplitude and phase features, respectively. $\bm{f}$ and $\bm{q}$ are the hidden vector from the masked features and the quantized feature as in figure, respectively. This example shows the case of two channels. In this approach, the quantizer converts concatenated vectors $[{X_1}^{\rm amplitude};{X_2}^{\rm amplitude};{X_2}^{\rm amplitude};{X_2}^{\rm phase}]$ to quantized vector $\bm{q}$. The quantizer consists of a single linear layer.

\begin{figure*}[htb]
	\begin{minipage}[b]{0.5\linewidth}
		\centering
		\includegraphics[scale=0.29]{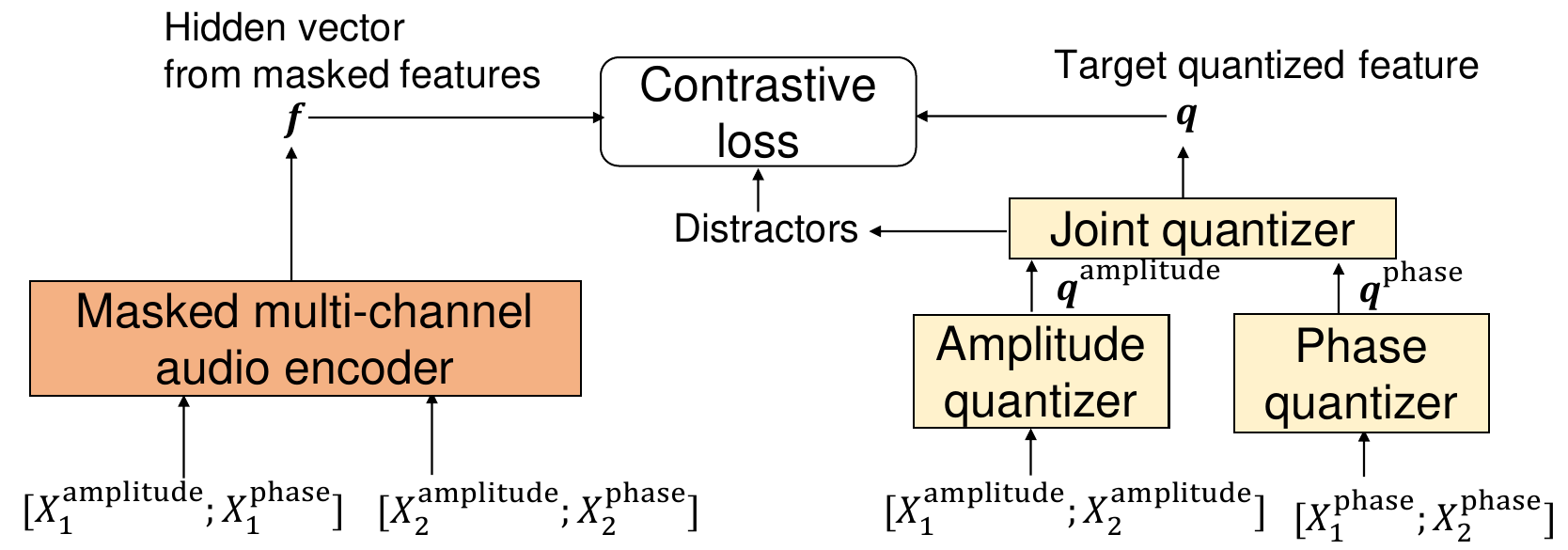}
		\caption{Feature-wise quantization.}
		\label{fig:p2}
	\end{minipage}
	\begin{minipage}[b]{0.5\linewidth}
		\centering
		\includegraphics[scale=0.29]{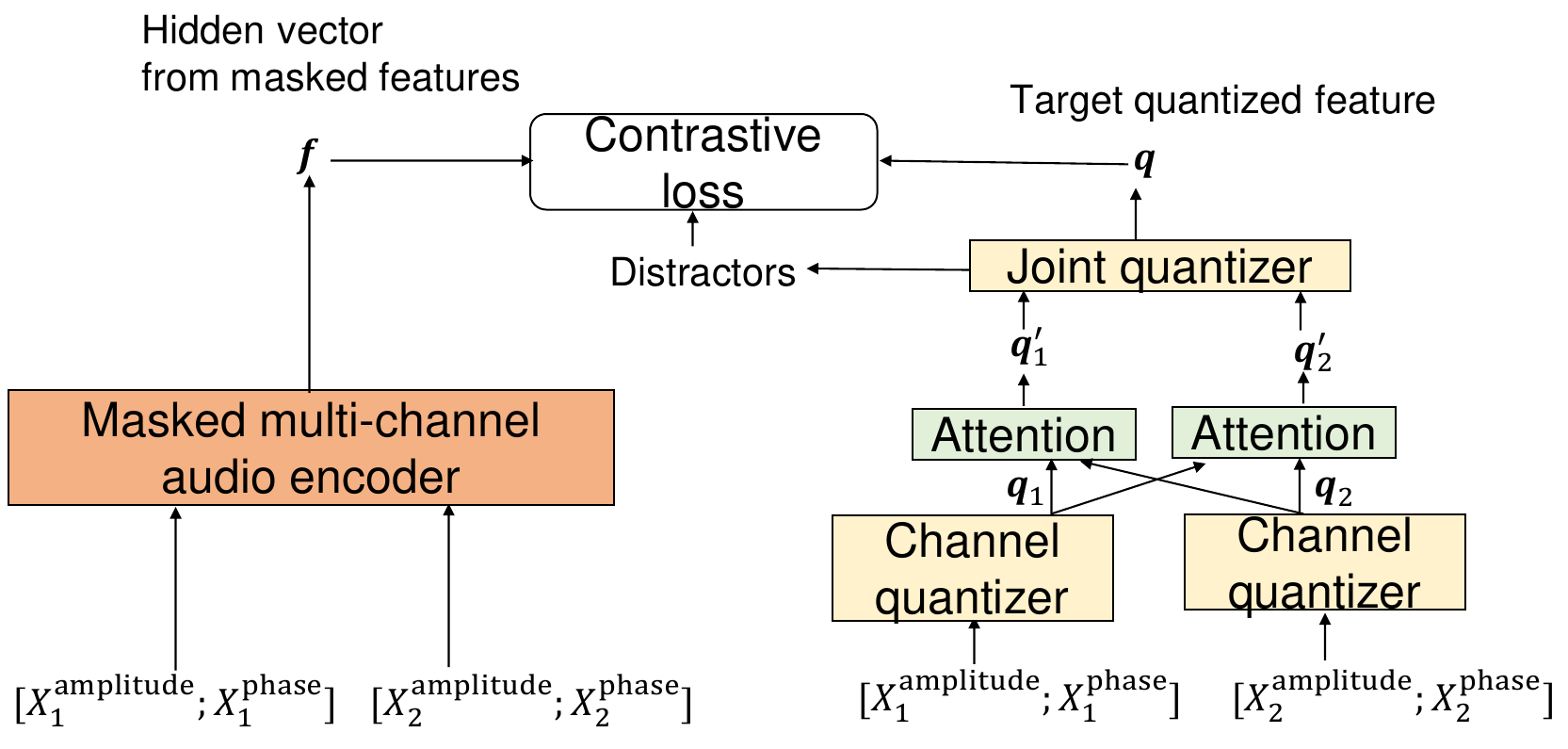}
		\caption{Channel-wise quantization.}
		\label{fig:p3}
	\end{minipage}	
\end{figure*}

\subsection{Feature-wise quantization}
\label{proposed1}
Figure~\ref{fig:p2} shows \mbox{feature-wise} quantization in the case of two channels. In this method, the quantization module consists of amplitude, phase and joint quantizers. The amplitude quantizer converts amplitude features $[{X_1}^{\rm amplitude};{X_2}^{\rm amplitude}]$ to quantized amplitude features $\bm{q}^{\rm amplitude}$. The phase quantizer converts phase features $[{X_1}^{\rm phase};{X_2}^{\rm phase}]$ to quantized phase features $\bm{q}^{\rm phase}$. 
The joint quantizer obtains the two vectors $[\bm{q}^{\rm amplitude};\bm{q}^{\rm phase}]$ from the two quantizers and converts them to quantized feature $\bm{q}$. All quantizers consist of linear layers. 

\subsection{Channel-wise quantization}
\label{proposed2}
Figure~\ref{fig:p3} shows \mbox{channel-wise} quantization. In this method, the quantization module consists of channel quantizers, the attention and the joint quantizer. The channel quantizers convert the amplitude and phase features $[{X_c}^{\rm amplitude};{X_c}^{\rm phase}]$ from each input channel to \mbox{channel-wise} quantized features ${\bm{q}_c}$. To capture the contextual relationship across channels, the attention was calculated.
For instance, the attention for the first channel in the case of two channels is calculated as
\begin{equation}
{\bm a}_1 = {\rm Softmax}(\bm{w}^T {\rm Tanh}(U X_1 + H X_2+\bm{b})),
\label{eq:attention}
\end{equation}
where $X_1=[{X_1}^{\rm amplitude};{X_1}^{\rm phase}]$, $X_2=[ {X_2}^{\rm amplitude}; {X_2}^{\rm phase}]$ and $\bm{w}$, $U$, $H$ and $\bm{b}$ are model parameters. The attention is used to calculate weighted quantized vector ${\bm{q}_c}^{'}$. The joint quantizer converts the weighted quantized features from each channel quantizer to quantized vector $\bm{q}$.

\section{Experiments}
\label{sec:experiment}

\subsection{Data preparation}
\label{sec:corpus}
\begin{description}
\item[Far-field in-house dataset]
As the experimental dataset, we use the \mbox{far-field in-house} dataset, which consists of 104.3 hours of transcribed Japanese speech. The speech is recorded by a \mbox{two-channel} linear microphone array with
an \mbox{inter-microphone} spacing of 8 mm. The amounts of training and test data are 102 and 2.3 hours, respectively. For \mbox{pre-training} and \mbox{fine-tuning}, we use the training set, and for the evaluation, we use the test set. For this dataset, we report the character error rate (CER).

\item[CHiME-4 dataset]
We also use the \mbox{CHiME-4} dataset to evaluate our proposed method. Speech in English is recorded by six microphones. For efficient training, we pick up the first and sixth microphone channels as input signals. In this experiment, we use training and evaluation sets on real data. For \mbox{pre-training} and \mbox{fine-tuning}, we use the training set. For evaluation, we use the eval set. For this dataset, we report the word error rate (WER) and CER.
\end{description}

\subsection{Model details}
\label{sec:archtec}
We next describe the architecture of the \mbox{multi-channel} neural transducer. We use eight conformer layers and a unidirectional long \mbox{short-term} memory (LSTM) layer with 256 hidden nodes for the \mbox{multi-channel} audio and label encoders, respectively. Table~\ref{tab:conformer} shows the parameters of the Conformer \cite{conformer} encoder model. The parameter size of the \mbox{multi-channel} audio encoder is 15.0 (M). The joint network obtains \mbox{512-dimensional} vectors from audio and label encoders, and outputs \mbox{256-dimensional} vector with Tanh activation. Finally, softmax outputs logits.

\begin{table}[htb]
	\centering	
	\caption{\mbox{Multi-channel} Conformer encoder architecture.}
	\begin{tabular}{c|l} \hline
		Parameter & Value \\ \hline 		
		Number of layers & 8 \\ \hline 	
		Number of channels & 2 \\ \hline 															
		Number of heads & 8 \\ \hline 							
		Head dimension & 32 \\ \hline 							
		Kernel size & 7 \\ \hline 									
		Number of hidden nodes & 256 \\ \hline 							
		\shortstack{Position-wise feed-forward \\ dimension} & 512 \\ \hline 
	\end{tabular}
	\label{tab:conformer}
\end{table}

For the amplitude feature, we use the \mbox{log-STFT} square magnitude. For the phase feature, 
we use cosine interchannel phase differences (cosIPD) and sinIPD \cite{cos_ipd}. The features are extracted every 10 ms with a window size of 25 ms from audio samples. We set the FFT size as 512.

In the wav2vec 2.0 \mbox{pre-training}, we train the \mbox{multi-channel} audio encoder by minimizing the contrastive loss. We mask 50\% of the time steps and set the number of distractors as 100. The distractors are uniformly sampled from other masked time steps of the same utterance.

For \mbox{fine-tuning}, we train the \mbox{multi-channel} neural transducer by minimizing the \mbox{RNN--T} loss. As the baseline system, we use the \mbox{multi-channel} neural transducer without any \mbox{pre-training}. For the \mbox{far-field in-house} dataset, the model outputs 715 characters and a blank token. For the \mbox{CHiME-4} dataset, the model outputs 26 lower-case alphabet characters, three special tokens (apostrophe, period and whitespace) and a blank token. In addition, gradient clipping is applied with a value of 5 to avoid an exploding gradient. We apply SpecAugment \cite{specaugment} to improve robustness. For the training of all models, we use the Transformer learning schedule \cite{transformer}. We also use the Adam optimizer \cite{adam}, setting ${\beta}_1=0.9$, ${\beta}_2=0.98$ and $\epsilon=10$. All networks are implemented using Pytorch \cite{torch}.

\subsection{Results}
\label{sec:results}

\subsubsection{Far-field in-house dataset}
\label{sec:far}
\begin{table}[htb]
	\begin{center}
		\caption{Results of feature-wise quantization for \mbox{far-field in-house} dataset. Results are given as relative character error rate reduction (CERR) [\%]. A positive value indicates an improvement.}
		\scalebox{0.9}{
			\begin{tabular}{llllll} \hline
				ID & pre-training & quantization & \shortstack{amplitude \\ quantizer \\ activation} & \shortstack{phase \\ quantizer \\ activation} & CERR (\%) \\ \hline
				exp0 & - & - & - & - & 0 \\ \hline		
				exp1 &\checkmark & feature & Swish & Swish & 58.1 \\											
				exp2 &\checkmark & feature & Swish & ReLU & 60.5 \\
				exp3 &\checkmark & joint & - & - & 62.1 \\			
				exp4 &\checkmark & feature & Swish & None & ${\bf 66}$ \\ \hline			
		\end{tabular}}
		\label{tab:result_1}
	\end{center}	
\end{table}

Table~\ref{tab:result_1} shows the results of \mbox{feature-wise} quantization for the \mbox{far-field in-house dataset}.
In this experiment, we investigate the effect of the activation function for the amplitude and phase quantizers. Compared with the result of the model without any \mbox{pre-training} (exp0), we observe an improvement for all \mbox{pre-training methods} (exp1, exp2, exp3, exp4). In addition, the amount of improvement depends on the activation function (exp1, exp2, exp4). We observe a 66\% relative reduction in CER using the quantization method employing the amplitude quantizer with Swish activation \cite{swish} and the phase quantizer without any activation (exp4). The CER of the method was lower than that for joint quantization (exp3). 

\begin{table}[htb]
	\begin{center}
		\caption{Results of channel-wise quantization for \mbox{far-field in-house} dataset. Results are given as relative character error rate reduction (CERR) [\%]. A positive value indicates an improvement.}
		\begin{tabular}{l|lll} \hline
			ID & pre-training & quantization & CERR (\%) \\ \hline
			exp0 & - & - & 0 \\ \hline		
			exp3 & \checkmark & joint & 62.1 \\
			exp5 & \checkmark & channel & 49.1 \\ \hline			
		\end{tabular}
		\label{tab:result_2}
	\end{center}	
\end{table}

Table~\ref{tab:result_2} shows the results of \mbox{channel-wise} quantization. Compared with the result of the model without any \mbox{pre-training} (exp0), the quantization method also reduced CER (exp5). The improvement was greater for joint quantization (exp3) than for \mbox{channel-wise} quantization (exp5).

\subsubsection{CHiME-4 dataset}
\label{sec:chime4}
Table~\ref{tab:result_3} shows the results of \mbox{feature-wise} quantization for the \mbox{CHiME-4} dataset. 
Compared with the result of the model without any \mbox{pre-training} (expA), we observe an improvement for all \mbox{pre-training methods} (expB, expC, expD, expE), the same as that for the \mbox{far-field in-house} dataset. We observe a 2.4\% relative reduction in WER using the quantization method employing the amplitude quantizer with Swish activation and the phase quantizer with Swish activation (expB). We observe a 4.2\% relative reduction in CER using the quantization method employing the amplitude quantizer with Swish activation and the phase quantizer without any activation (expE). Comparing the improvement of CER the for \mbox{far-field in-house} dataset, the improvement of CER for the \mbox{CHiME-4} dataset was small. We concluded that this is caused by the amount of training data in the \mbox{CHiME-4} dataset being smaller than that in the \mbox{far-field in-house} dataset.

\begin{table}[htb]
	\begin{center}
		\caption{Results of feature-wise quantization for \mbox{CHiME-4} dataset. Results are given as relative character error rate reduction (CERR) [\%] and relative word error rate reduction (WERR) [\%]. A positive value indicates an improvement.}
		\scalebox{0.7}{
			\begin{tabular}{lllllll} \hline
				ID & pre-training & quantization & \shortstack{amplitude \\ quantizer \\ activation} & \shortstack{phase \\ quantizer \\ activation} & CERR (\%) & WERR (\%) \\ \hline
				expA & - & - & - & - & 0 & 0 \\ \hline		
				expB &\checkmark & feature & Swish & Swish & 3.6 & {\bf 2.4} \\ 	
				expC &\checkmark & feature & Swish & ReLU & 2.6 & 1.7 \\
				expD &\checkmark & joint & - & - & 3.2 & 1.4 \\			
				expE &\checkmark & feature & Swish & None & {\bf 4.2} & 0.1 \\ \hline			
		\end{tabular}
	}
		\label{tab:result_3}
	\end{center}	
\end{table}

\subsection{Analysis of hidden vectors}
\label{sec:analyze_hidden}
\begin{figure}[htb]
	\centering
	\includegraphics[scale=0.16]{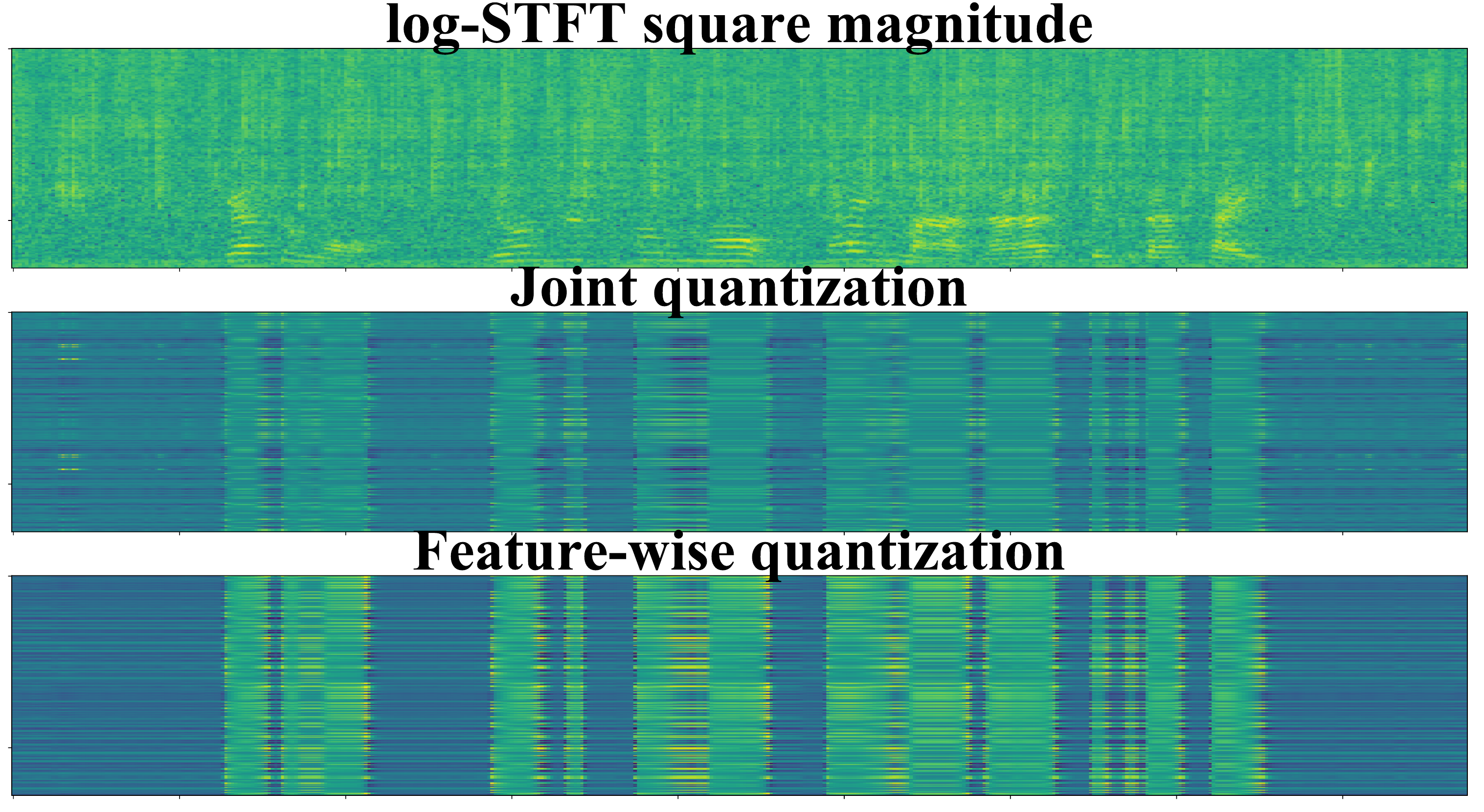}	
	\caption{Analysis of hidden vectors after self-supervised learning.}
	\label{fig:ssl}
\end{figure}
We next analyze the hidden vectors from the \mbox{multi-channel} audio encoder after \mbox{pre-training}. Figure~\ref{fig:ssl} shows hidden vectors from \mbox{multi-channel} audio encoders trained by different quantization methods for the \mbox{far-field in-house dataset}. The upper figure shows the \mbox{log-STFT} square magnitude. The middle figure shows the hidden vector from the \mbox{multi-channel} audio encoder trained by joint quantization (exp3). The lower figure shows the hidden vector from the \mbox{multi-channel} audio encoder trained by \mbox{feature-wise} quantization (exp4). By comparing the hidden vectors, we observe a clearer contrast between the speech and noise sections for \mbox{feature-wise} quantization. This result suggests that the \mbox{multi-channel} audio encoder trained by \mbox{feature-wise} quantization learns the latent representation better than the \mbox{multi-channel} audio encoder trained by joint quantization in terms of noise robustness.

\section{Conclusion}
\label{sec:conclution}
In this work, we trained a \mbox{multi-channel} neural transducer based on wav2vec 2.0 \mbox{pre-training}. 
For the training, we explored three quantization methods: joint quantization, \mbox{feature-wise} quantization and \mbox{channel-wise quantization}. 
We reported the results of experiments using the \mbox{far-field in-house} and public datasets. We experimentally showed that the \mbox{feature-wise quantization} method had the best performance. We observed 66\% and 4.2\% relative reductions in CER compared with the model without any \mbox{pre-training} for the \mbox{far-field in-house} and \mbox{CHiME-4} datasets, respectively.


\begin{thebibliography}{99} 
	
	\bibitem{las}
	W. Chan, N. Jaitly, Q. Le and O. Vinyals,
	``Listen, attend and spell: A neural network for large vocabulary conversational speech recognition,''
	in \textit{Proc. ICASSP}, 2016.		

	\bibitem{ctc}
	A. Graves, S. Fern\'andez, F. Gomez and J. Schmidhuber,
	``Connectionist temporal classification: Labelling unsegmented sequence data with recurrent neural networks,''
	in \textit{Proc. ICML}, 2006.

	\bibitem{rnnt}
	A. Graves,
	``Sequence transduction with recurrent neural networks,''
	\textit{arXiv preprint arXiv:1211.3711}, 2012.
	
	
	\bibitem{w2v}
	A. Baevski, H. Zhou, A. Mohamed and M. Auli,
	``wav2vec 2.0: A framework for self-supervised learning of speech representations,'' 
	\textit{arXiv preprint arXiv:2006.11477}, 2020.
	
	\bibitem{bert}
	J. Devlin, M.-W. Chang, K. Lee and K. Toutanova,
	``Bert: Pre-training of deep bidirectional transformers for language understanding,''
	arXiv, abs/1810.04805, 2018.	
	
	\bibitem{mel2vec}
	Y. Zhang, J. Qin, D. S. Park, W. Han, C.-C. Chiu, R. Pang, Q. V. Le and Y. Wu,
	``Pushing the limits of semi-supervised learning for automatic speech recognition,''
	\textit{arXiv preprint arXiv:2010.10504}, 2020.	

	\bibitem{w2v_c}
	S. Sadhu, D. He, C.-W. Huang, S. H. Mallidi, M. Wu, A. Rastrow, A. Stolcke, J. Droppo and R. Maas,
	``Wav2vec-C: A self-supervised model for speech representation learning,''
	 in \textit{Proc. INTERSPEECH}, 2021.		

	\bibitem{beam}
	R. Haeb-Umbach, J. Heymann, L. Drude, S. Watanabe, M. Delcroix and T. Nakatani,
	``Far-field automatic speech recognition,''
	\textit{Proc. IEEE}, 2020.	
	
	\bibitem{chime4}E. Vincent, S. Watanabe, A. A. Nugraha, J. Barker and R. Marxer,
	``An analysis of environment, microphone and data simulation mismatches in robust speech recognition,'' \textit{Computer Speech \& Language}, vol. 46, pp. 535-557, 2017.


	\bibitem{multi_las1}	
	T. Ochiai, S. Watanabe, T. Hori and J. R. Hershey,
	 ``Multichannel end-to-end speech recognition,'' \textit{arXiv preprint arXiv:1703.04783}, 2017.
	 
	 
	\bibitem{multi_las2}	
	X. Chang, W. Zhang, Y. Qian, J. Le Roux and S. Watanabe,
	``MIMO-SPEECH: End-to-end multi-channel multi-speaker speech recognition,'' in \textit{Proc. ASRU}, 2019.
	
	\bibitem{multi_las3}
	A. S. Subramanian, X. Wang, S. Watanabe, T. Taniguchi, D. Tran and Y. Fujita,
	``An investigation of end-to-end multichannel speech recognition for reverberant and mismatch conditions,'' \textit{arXiv preprint arXiv:1904.09049}, 2019.		
	
	\bibitem{multi_las4}		
	F. Chang, M. Radfar, A. Mouchtaris, B. King and S. Kunzmann,
	 ``End-to-end multi-channel Transformer for speech recognition,''	
	in \textit{Proc. ICASSP}, 2021.	 
	
	\bibitem{multi_tt}
	F. Chang, M. Radfar, A. Mouchtaris and M. Omologo,
	``Multi-channel Transformer Transducer for speech recognition,'' in \textit{Proc. INTERSPEECH}, 2021.
	
	\bibitem{bf_g}
	B. Li, T. N. Sainath, R. J. Weiss, K. W. Wilson and M. Bacchiani,
	``Neural network adaptive beamforming for robust multichannel speech recognition,'' in \textit{Proc. INTERSPEECH}, 2016.			
	

	\bibitem{t_t1}C. F. Yeh, J. Mahadeokar, K. Kalgaonkar, Y. Wang, D. Le, M. Jain, K. Schubert, C. Fuegen and M. L. Seltzer,
``Transformer--Transducer: End-to-end speech recognition with self-attention,''
in \textit{Proc. INTERSPEECH}, 2020.

\bibitem{t_t2}Q. Zhang, H. Lu, H. Sak, A. Tripathi, E. McDermott, S. Koo and S. Kumar,
``Transformer Transducer: A streamable speech recognition model with Transformer encoders and RNN--T loss,''
in \textit{Proc. INTERSPEECH}, 2020.

	\bibitem{conformer}
A. Gulati, J. Qin, C.-C. Chiu, N. Parmar, Y. Zhang, J. Yu, W. Han, S. Wang, Z. Zhang, Y. Wu and R. Pang,
``Conformer: Convolution-augmented transformer for speech recognition,''in \textit{Proc. INTERSPEECH}, 2020.
	
	\bibitem{mask_bf}
	J. Heymann, L. Drude and R. Haeb-Umbach,
	``Neural network based spectral mask estimation for acoustic beamforming,'' in \textit{Proc. ICASSP}, 2016.
	
	\bibitem{mask_bf2}
	H. Erdogan, J. Hershey, S. Watanabe, M. Mandel, and J. Le Roux,
	``Improved MVDR beamforming using single-channel mask prediction networks,'' in \textit{Proc. INTERSPEECH}, 2016.
	
	\bibitem{frame}
	T. Higuchi, K. Kinoshita, N. Ito, S. Karita and T. Nakatani,
	``Frame-by-frame closed-form update for mask-based adaptive MVDR beamforming,'' in \textit{Proc. ICASSP}, 2018.
	

	

	\bibitem{transformer} A. Vaswani, N. Shazeer, N. Parmar, J. Uszkoreit, L. Jones, A. N.
	Gomez, L. Kaiser and I. Polosukhin,
	``Attention is all you need,''
	 in \textit{Proc. NIPS}, 2017.
	
	
	\bibitem{cos_ipd}
	Z. Wang, J. Le Roux and J. R. Hershey,
	 ``Multi-channel deep clustering: Discriminative spectral and spatial embeddings for speaker-independent speech separation,'' in \textit{Proc. ICASSP}, 2018.
	 	 
	

	
	

	\bibitem{specaugment}
	D. S. Park, W. Chan, Y. Zhang, C. Chiu, B. Zoph, E. D. Cubuk and Q. V. Le,
	``SpecAugment: A simple data augmentation method for automatic speech recognition,''
	in \textit{Proc. INTERSPEECH}, 2019.	
	
	\bibitem{adam}
	D. P. Kingma and J. Ba,
	``Adam: A method for stochastic optimization,''
	in \textit{Proc. ICLR}, 2015.
	
	\bibitem{torch}
	A. Paszke, S. Gross, F. Massa, A. Lerer, J. Bradbury, G. Chanan, T. Killeen, Z. Lin, N. Gimelshein,
	L. Antiga, A. Desmaison, A. K\"opf, E. Yang, Z. DeVito, M. Raison, A. Tejani, S. Chilamkurthy, B. Steiner, L. Fang, J. Bai and S. Chintala,
	``PyTorch: An imperative style, high-performance deep learning library,''
	in \textit{Proc. NeurIPS}, 2019.
	
	\bibitem{swish}
P. Ramachandran, B. Zoph and Q. V Le,
``Searching for activation functions,''
\textit{arXiv preprint arXiv:1710.05941}, 2017.		 		
	
	
	

\end{thebibliography}
\end{document}